\documentclass[journal]{IEEEtran}
\usepackage{graphicx}
\usepackage{multirow}
\usepackage[table]{xcolor}
\definecolor{lightblue}{HTML}{AFFCFF}
\definecolor{lightyellow}{HTML}{FFFFC7}
\definecolor{lightgreen}{HTML}{DFFAD4}
\definecolor{lightpink}{HTML}{FAE2E1}
\definecolor{lightpurple}{HTML}{D8A9E6}
\usepackage{makecell}
\usepackage{arydshln}
\usepackage{algorithm}%
\usepackage{algorithmicx}%
\usepackage{algpseudocode}%
\usepackage{amsthm,amsmath,amssymb}
\usepackage{mathrsfs}
\usepackage{makecell}
\usepackage{enumitem}
\usepackage{caption}
\usepackage{wrapfig}
\usepackage{pifont}
\usepackage{subcaption}
\usepackage{enumitem}
\usepackage{marvosym}

\newtheorem{theorem}{Theorem}[section]

\newtheorem{definition}[theorem]{Definition}

\ifCLASSINFOpdf
\else
\fi
\begin{document}
\title{HateDebias: On the Diversity and Variability of Hate Speech Debiasing}



\author{\IEEEauthorblockN{
Hongyan Wu, Zhengming Chen, Zijian Li, Nankai Lin\textsuperscript{\Letter}, Lianxi Wang, Shengyi Jiang and Aimin Yang
}

\thanks{
Corresponding author: Nankai Lin (email: neakail@outlook.com).}
\thanks{Hongyan Wu is with the College of Computer, National University of Defense Technology, Hunan, China.}
\thanks{Zhengming Chen is with College of Science, Shantou University, Guangdong, China.}
\thanks{Zijian Li is with Mohamed bin Zayed University of Artificial Intelligence, Abu Dhabi, United Arab Emirates.}
\thanks{Nankai Lin, Lianxi Wang and Shengyi Jiang are with the School of Information Science and Technology, Guangdong University of Foreign Studies, Guangdong, China.}
\thanks{Aimin Yang is with School of Computer Science, Guangdong University of Technology, Guangdong, China.}
}

%



\IEEEtitleabstractindextext{%
\begin{abstract}
Hate speech frequently appears on social media platforms and urgently needs to be effectively controlled. Alleviating the bias caused by hate speech can help resolve various ethical issues. Although existing research has constructed several datasets for hate speech detection, these datasets seldom consider the diversity and variability of bias, making them far from real-world scenarios. To fill this gap, we propose a benchmark \textbf{HateDebias} to analyze the fairness of models under dynamically evolving environments. Specifically, to meet the diversity of biases, we collect hate speech data with different types of biases from real-world scenarios. To further simulate the variability in the real-world scenarios(i.e., the changing of bias attributes in datasets), we construct a dataset to follow the continuous learning setting and evaluate the detection accuracy of models on the \textbf{HateDebias}, where performance degradation indicates a significant bias toward a specific attribute. To provide a potential direction, we further propose a continual debiasing framework tailored to dynamic bias in real-world scenarios, integrating memory replay and bias information regularization to ensure the fairness of the model. Experiment results on the HateDebias benchmark reveal that our methods achieve improved performance in mitigating dynamic biases in real-world scenarios, highlighting the practicality in real-world applications.
\end{abstract}

\begin{IEEEkeywords}
Hate Speech Debiasing, Continuous Learning, Diversity, Variability, Benchmark, Continuous Debiasing.
\end{IEEEkeywords}}

\maketitle

\IEEEdisplaynontitleabstractindextext

%
\IEEEpeerreviewmaketitle

\section{Introduction}
\label{intro}

Hate speech \cite{ghosh-etal-2023-cosyn, yu-etal-2023-fine} is frequent on social media, which has become a critical social problem and is desired to solve. One mainstream solution is to employ the technique of machine learning to detect and further mitigate hate speech \cite{nguyen-etal-2023-towards, roy-etal-2023-probing, park-etal-2023-uncovering}, hence a safe and inclusive environment of the internet can be protected by removing these hate speeches. {To advance algorithms for hate speech detection, it is important to provide benchmarks that are close to real-world scenarios.}  

{The hate speech detection task aims to identify whether a given speech is hate speech or not.} There are different types of benchmarks to evaluate the hate speech detection algorithm, which can be summarized in Table \ref{table1}. {Previously, ETHOS \cite{mollas2022ethos} collected data from YouTube and Reddit comments, which is designed to recognize the target of hate speech.} {The Measuring Hate Speech corpus \cite{sachdeva-etal-2022-measuring} is devised to identify the target community and classify a sentence as hate speech or not}. In summary, these datasets are devised to recognize the target of hate speech. Recently, researchers have aimed to address the problem of debiasing and propose different datasets. Specifically, RaceBiasHate \cite{davidson-etal-2019-racial} uses the Twitter data annotated for hate speech and abusive language to address the racial bias. Geo-CulturalBiasHate \cite{tonneau2024languages} evaluates cultural bias in hate speech datasets by leveraging two interrelated cultural proxies: language and geography. Furthermore, the MultilingualBiasHate corpus \cite{huang-etal-2020-multilingual} further considers different types of hate speech, consisting of age, country, gender and race. In summary, existing datasets usually focus on the target of hate speech and seldom consider different types of biases.

\begin{figure*}[t]
  \centering
  \includegraphics[width=0.7\textwidth]{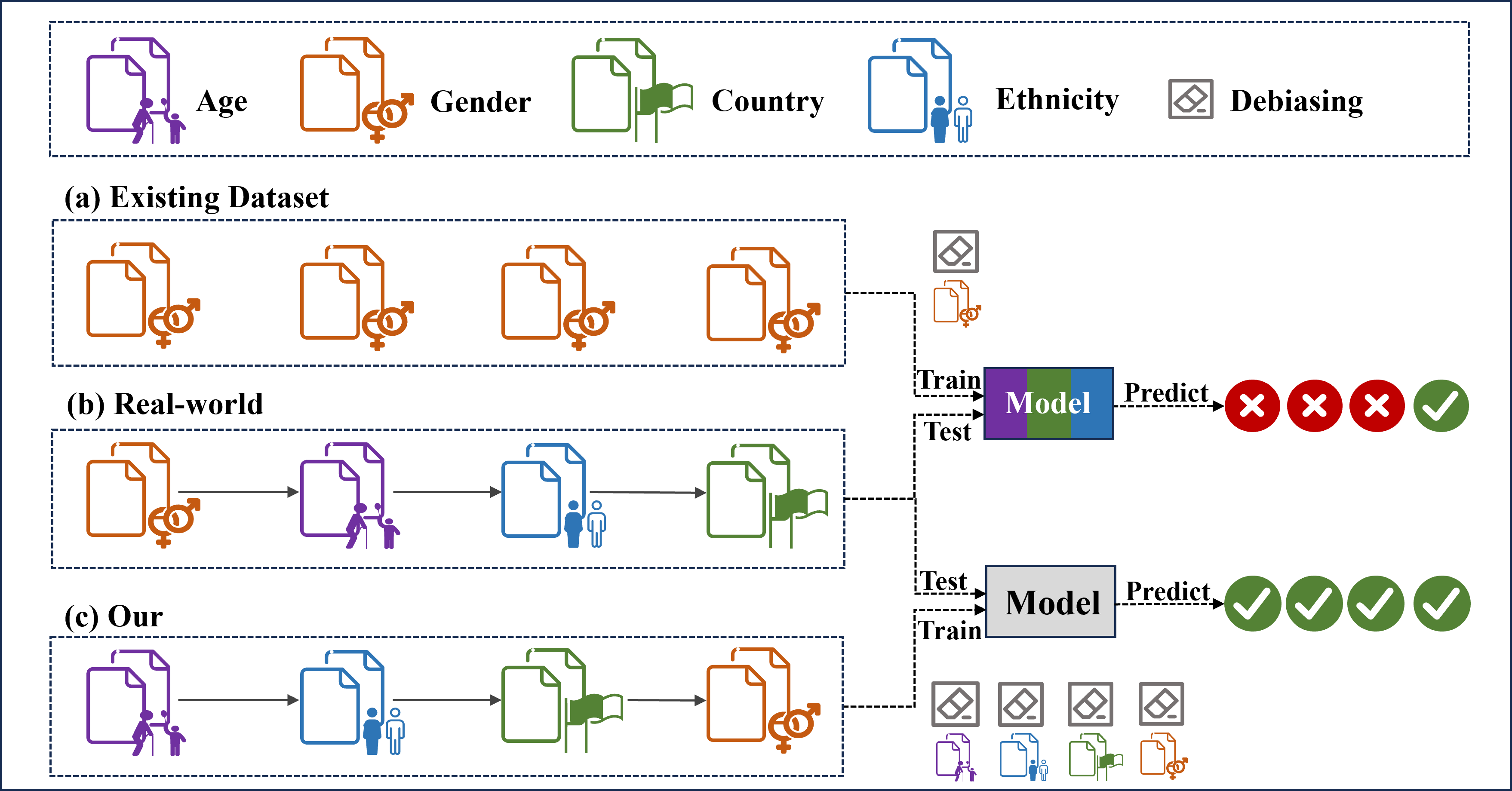}
  \caption{Comparison of existing datasets and our dataset with real-world scenarios. Figure (a) shows that existing datasets typically contain a single type of bias, which may lead to suboptimal performance of models when dealing with real-world scenarios involving multiple and varying biases. Figure (b) indicates that in real-world scenarios, there are multi-types and continuous-varying biases. Figure (c) demonstrates our constructed HateDebias dataset, which simulates the continuous emergence of various types of hate speech biases, making the dataset distribution more similar to real-world scenarios and thereby improving model performance in practical applications.} 
  \label{fig:motivation}
\end{figure*}

Considering the multiple types of biases of hate speech is essential to developing hate speech debiasing models, as it prevents models from ignoring different biases during training. Although these datasets have boosted the development of hate speech detection algorithms to some extent, several algorithms trained on these datasets can hardly achieve ideal performance in practice since they are still far from real-world scenarios. As shown in Figure \ref{fig:motivation} (a), since existing datasets contain a single type of bias, the models trained on these datasets may achieve suboptimal performance on real-world scenarios with multiple and time-varying biases in Figure \ref{fig:motivation} (b). To address this challenge, it is desired to build a benchmark that is close to a real-world application. As shown in Figure \ref{fig:motivation} (c), when simulating how the multiple types of hate speech occur continuously, we can build a dataset with a similar distribution as the real-world scenarios, so the models may achieve ideal performance.

\begin{table}
\caption{Comparisons of different datasets. BI: Biased Information; D: Diversity; V: Variability.}
\centering
\begin{tabular}{cccc}
\hline
Dataset & BI & D & V \\
\hline
ETHOS \cite{mollas2022ethos} & \ding{56} & \ding{56} & \ding{56} \\
Measuring Hate Speech \cite{sachdeva-etal-2022-measuring} & \ding{56} & \ding{56} & \ding{56} \\
Geo-CulturalBiasHate \cite{tonneau2024languages} & \ding{52} & \ding{56} & \ding{56} \\
RaceBiasHate \cite{davidson-etal-2019-racial} & \ding{52} & \ding{56} & \ding{56} \\
MultilingualBiasHate \cite{huang-etal-2020-multilingual} & \ding{52} & \ding{52} & \ding{56} \\
\hdashline
HateDebias & \ding{52} & \ding{52} & \ding{52} \\
\hline
\end{tabular}
\label{table1}
\end{table}

To achieve this, we first formalize the hate speech scenarios with multiple types and dynamically varying biases as continuous debiasing tasks. {Our goal is to alleviate hate speech biases toward the specific attribute in each sub-dataset while maintaining promising performance on hate speech detection tasks.} Sequentially, we construct a real-world dataset, HateDebias, to serve as a benchmark for these continuous debiasing tasks. It consists of 23276 hate speech texts with 4 types of bias attributes, including age, country, gender, and ethnicity. Specifically, we simulate the emergence of various new biases at different stages based on the \cite{huang-etal-2020-multilingual} and \cite{wang-etal-2019-sentence}, such that each type of bias corresponds to a specific sub-dataset in HateDebias. As shown in Table \ref{table1}, one can see that our dataset offers greater diversity and variability compared to existing hate speech detection datasets. Moreover, we propose a continual-learning-based framework, leveraging a bias information regularization (BIR) strategy and a memory replay (MR) strategy to better understand and tackle biases in the training process.

\textbf{Contribution.} (1) We explicitly formalize the continuous debiasing task that tackles the continuous debiasing scenario with multi-types and continuous-varying bias. (2) We construct a real-world dataset, called HateDebias, where any sequence of sub-dataset in HateDebias corresponds to continuous-varying bias. (3) We propose a simple yet effective framework for continuous debiasing tasks, which provides a potential direction to improve continuous debiasing tasks.

\section{Related Work}

In this section, we briefly review the related literature, including datasets of hate speech detection, static debiasing, continuous debiasing, and continual learning.

\begin{figure*}[htbp]
  \centering
  \includegraphics[width=0.7\textwidth]{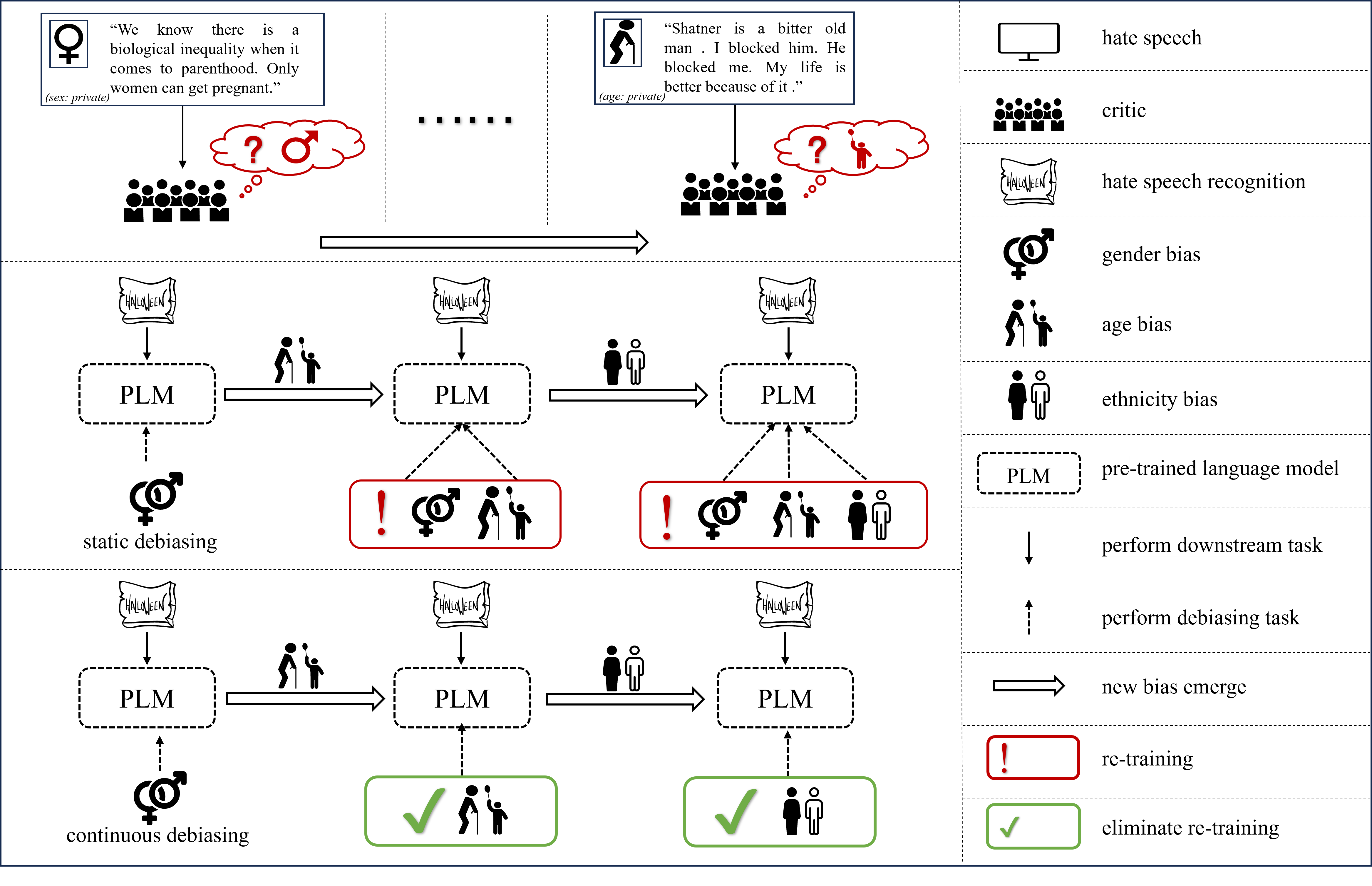}
  \caption{An illustration of stastic debiasing task and continuous debiasing task. In the static debiasing task, the emergence of a new bias requires a complete re-debiasing of all existing biases, resulting in an exponential increase in both training time and costs.  Conversely, in the continuous debiasing task, the emergence of a new bias requires only the fine-tuning of this specific bias, eliminating the need for re-training.} 
  \label{fig:1} 
\end{figure*}

\paragraph{Datasets of Hata Speech Detection.} Hate speech detection has been a focal point in natural language processing research, leading to the creation of several datasets aimed at improving the detection capabilities of various models. These datasets vary significantly in their sources, annotations, and the types of biases they address, providing a broad spectrum of resources for developing and evaluating hate speech detection algorithms. One of the earlier datasets is ETHOS \cite{mollas2022ethos}, which is built from YouTube and Reddit comments validated through a crowdsourcing platform. ETHOS is designed to recognize the target of hate speech, focusing primarily on detecting who the hate speech is directed towards. Another significant dataset is the Measuring Hate Speech corpus \cite{sachdeva-etal-2022-measuring}, which aims to identify the target community and classify whether a sentence contains hate speech. This dataset emphasizes the categorization of the hate speech target, adding a layer of granularity to hate speech detection. Recent advancements have seen the development of datasets addressing specific types of biases. For example, RaceBiasHate \cite{davidson-etal-2019-racial} uses Twitter data annotated for hate speech and abusive language to tackle racial bias. This dataset is crucial for understanding and mitigating racial biases in hate speech detection systems. Similarly, Geo-CulturalBiasHate \cite{tonneau2024languages} evaluates cultural bias in hate speech datasets by leveraging two interrelated cultural proxies: language and geography. This dataset underscores the importance of considering geographic and cultural contexts in hate speech detection. The MultilingualBiasHate corpus \cite{huang-etal-2020-multilingual} expands the scope by including different types of hate speech related to age, country, gender, and race. This multilingual approach provides a comprehensive resource for training and evaluating models on a variety of hate speech scenarios across different languages.

\paragraph{Static Debiasing.}
In the field of Natural Language Processing (NLP), bias mitigation strategies fall into two primary categories. The first involves training models to reduce the influence of sensitive attributes during fine-tuning \cite{basu2023efficient,lu2024debiasing}, using techniques like projection-based \cite{ravfogel2022linear}, adversarial \cite{han-etal-2021-diverse}, or contrastive objectives \cite{chi-etal-2022-conditional}. This approach requires data specifically annotated for the sensitive attribute, with the debiasing effectiveness dependent on the external corpus quality and the potential introduction of new biases. The second category, task-agnostic training, uses general corpora to address biases, such as by identifying and removing gender biases from encoded representations \cite{kaneko-bollegala-2021-debiasing} or enhancing model fairness through selective data exclusion \cite{NEURIPS2022_e94481b9}, increased dropout rates \cite{DBLP:journals/corr/abs-2010-06032}, or equalizing objectives \cite{guo-etal-2022-auto}. Additionally, methods like the introduction of fairness-triggering perturbations \cite{NEURIPS2022_de08b3ee} and equi-tuning for group equivariance in immutable models \cite{Basu_Sattigeri_Natesan} have been explored, focusing on minimizing differences between standard and debiased model features. 

\paragraph{Continuous Debiasing.}
Current research on debiasing word embeddings mainly addresses static biases and stereotypes, with efforts on pre-trained language models targeting known sensitive attributes \cite{lauscher-etal-2021-sustainable-modular,guo-etal-2022-auto}. However, unforeseen biases cannot be fully anticipated due to societal changes. Google's ML-fairness-gym framework aims to simulate continuous debiasing in machine learning, analyzing the impact of algorithms on information systems and human behavior, yet it does not extend to unstructured data, such as text \cite{10.1145/3351095.3372878}. To date, continuous debiasing in text processing remains unexplored.

\paragraph{Continual Learning.}
Continual learning involves updating models with new information while retaining previous knowledge, avoiding catastrophic forgetting \cite{10.1145/3543507.3583262}. This process entails maintaining generic internal representations for reuse across tasks and dynamically adjusting task-specific parts. Effective strategies include memory-based methods that use prior task exemplars \cite{riemer2018learning} or synthetic data \cite{9522984}, and regularization techniques that minimize changes in learned representations under a fixed architecture \cite{akyurek2021subspace}. For example, Li and Hoiem \cite{10.1109/TPAMI.2017.2773081} proposed "learning without forgetting" (LwF), a technique that introduces regularization to align the model’s output on the current data with that of the previously trained model. Knowledge distillation \cite{Cheraghian_2021_CVPR}, gradient-based \cite{9578703} and expansion-based approaches \cite{10.5555/3454287.3455512} also mitigate forgetting, each targeting the preservation and enhancement of model adaptability over time. Furthermore, some research endeavors have shifted attention toward continual-learning-based methodologies as potential solutions to address model fairness concerns. Churamani et al. \cite{churamani2022domain} introduced a pioneering application of domain-incremental learning as an effective approach for mitigating bias, specifically in the context of enhancing the fairness of facial expression recognition systems. However, there is no work focusing on the continuous debiasing task.

\section{Illustration of Continuous Debiasing and Static Debiasing}

{In our work, the term ``bias'' refers to the unfair predictions made by the model towards speech originating from certain groups \cite{guo-etal-2022-auto, bias}. For example, when analyzing the text of a man's speech, the model might incorrectly classify it as hate speech, influenced by the male attribute. The performance of a fair model in predicting hate speech remains consistent and unbiased, unaffected by the group attributes of the speech \cite{kleinberg2016inherent, hardt2016equality}.} Based on this definition of bias, the task of debiasing is traditionally categorized into two types: static debiasing and continuous debiasing.


{\textbf{Static Debiasing.}} The static debiasing is not sufficient for covering real-world scenarios. Traditionally, the static debiasing focused on adjusting specific biases, as illustrated in Fig. \ref{fig:1}. Static debiasing requires re-debiasing all biases whenever new ones appear, consuming considerable training time and resources. Such a cost makes its application impractical in real-world environments where new biases emerge continuously. For instance, if a model is initially trained to mitigate racial bias, and then gender bias becomes apparent, the entire model needs to be retrained to address both biases simultaneously. This retraining process can be highly inefficient and resource-intensive, especially as the number and types of biases continue to grow.

{\textbf{Continuous Debiasing.} In continuous debiasing, models in ever-changing environments are expected to adjust targetedly and dynamically to correct these continuously emerging biases, as shown at the bottom of Fig. \ref{fig:1}. The continuous debiasing allows for incremental updates to the model to address new biases as they arise without requiring retraining from scratch.} This task is significantly efficient and scalable, making it more suitable for real-world applications where the social and cultural context is continuously evolving.

Continuous debiasing plays a crucial role as it enables models to adapt to the dynamic nature of biases in the real world. However, despite its significance, there is a notable lack of benchmarks available for continuous debiasing tasks. Without proper benchmarks, it becomes difficult to gauge the effectiveness of new methods and to determine which ones are truly capable of handling the complex and evolving nature of biases in real-world scenarios. Therefore, in this paper, \textbf{our goal} is to propose a benchmark specifically designed for continuous debiasing tasks. We will define the continuous debiasing task and construct a comprehensive dataset that encompasses a wide range of biases and scenarios that mimic real-world variations. In addition, we develop a method that is tailored to work effectively with our proposed benchmark.


\section{Benchmark Continuous Debiasing Tasks}

In this section, we aim to benchmark the continuous debiasing task. The proposed HateDebias benchmark includes three aspects: task definition, dataset construction and practical framework. Specifically, we first explicitly define the continuous debiasing task that aims to continually update the model to adapt to new biases as they emerge (Sec. \ref{s4}). Secondly, we propose a real-world dataset in which various biases are introduced periodically, used as the benchmark for evaluating continuous debiasing tasks (Sec. \ref{s5}). Finally, we design a continual-learning-based framework incorporating techniques like bias regularization and memory replay to effectively debias over time, as a general solution for handling continuous debiasing tasks (Sec. \ref{s6}).

\subsection{Continuous Debiasing Task Definition}\label{s4}

{Continuous debiasing requires that models adaptively adjust in the attribute-incremental continual learning scenarios and incrementally address newly emerging biases over time. Specifically, for hate speech detection, the model initially is deployed to detect hate speech and mitigate \textit{racial bias} related to the user. In a dynamic environment, \textit{gender-biased} hate speech gradually occurs on the platform. In this case, the continuous debasing task aims to develop a model that can be adapted to mitigate racial and gender biases in an evolving environment without neglecting its original capability.} 

Supposing that $D_i$ is a dataset and $A_i$ represents the bias attribute corresponding to $D_i$, the continuous debias task is defined as follows:


\begin{definition}[Continuous Debiasing Task]
    {Given a hate speech dataset $\mathcal{D} = \{D_1, \cdots, D_n\}$ with its corresponding bias attribute $A_i \in \mathcal{A}$, where $\mathcal{A} = \{\text{Age},\text{Gender},\text{Country},\text{Ethnicity} \}$. The continuous debiasing task is a task sequence $\mathcal{T}_{\textit{debias}} = \{\operatorname{T_1}, \cdots, \operatorname{T_n}\}$, aiming to debias for the attribute in each sub-dataset $D_i$ sequentially.}
\end{definition}

It is worth noting that the order of bias attribute in $\mathcal{A}$ can be arbitrary, implying the diversity and variability of bias attribute for the dataset $\mathcal{D}$. Moreover, the goal of our work is not only to debias the hate speech bias embedded in the model but also to maintain performance in downstream tasks applied to the dataset $\mathcal{D}$, such as classification tasks.

There are two key characteristics of the continuous debiasing task: diversity and variability. We will properly define them to induce the construction of the benchmark dataset. \textbf{Diversity} aims to emphasize the multi-type bias attributes within the dataset $\mathcal{D}$, e.g., $|\mathcal{A}| \geq 3$ where $|\cdot|$ represents the dimension. On the other hand, \textbf{variability} represents the difference in bias attributes across various stages of the task. For instance, in a continuous debiasing task sequence $\mathcal{T}_{\textit{debias}} = \{\operatorname{T}_1, \cdots, \operatorname{T}_n\}$, $\operatorname{T}_i$  aims to debias the bias attribute $A_i$ while $\operatorname{T}_{i+1}$ addresses the bias attribute $A_j$, with ${A_i} \neq A_j$.


\subsection{Continuous Debiasing Dataset}\label{s5}

After a well-defined continuous debiasing task, this section will provide the detailed construction of our benchmark dataset, HateDebias, which can be used to evaluate the continuous debiasing task and accordingly encourage the development of related methods later. To better simulate scenarios that are close to the real-world environment, the HateDebias dataset is required to satisfy two conditions in our construction procedure: diversity and variability of biases.

\begin{table}
\tabcolsep=0.08cm
\centering
\caption{Label and four attributes distribution of the training set. The row represents datasets and the column shows the attributes and labels within these datasets, and the table values detail the distribution of each attribute and label. The term ``Dataset'' refers to the aggregate of all four subsets. 
}
\begin{tabular}{ccccccc}
\hline
\multirow{2}{*}{Attribute} & \multirow{2}{*}{Value} & \multirow{2}{*}{Dataset} & \multicolumn{4}{c}{Subset} \\
\cline{4-7}
& & & Age & Gender & Country & Ethnicity \\
\hline
\multirow{2}{*}{Age} & elder & 10561 & 2640 & 2639 & 2643 & 2639 \\
 & median  & 12715 & 3179 & 3180 & 3176 & 3180 \\
 \hline
\multirow{2}{*}{Gender} & male & 8850 & 2212 & 2213 & 2212 & 2213 \\
 & female & 14426 & 3607 & 3606 & 3607 & 3606 \\
 \hline
\multirow{2}{*}{Country} & non-US & 7418 & 1855 & 1854 & 1854 & 1855 \\
 & US & 15858 & 3964 & 3965 & 3965 & 3964 \\
 \hline
\multirow{2}{*}{Ethnicity} & non-white & 15480 & 3871 & 3871 & 3869 & 3869 \\
 & white & 7796 & 1948 & 1948 & 1950 & 1950 \\
 \hline
\multirow{2}{*}{Label} & non-hate & 17026 & 4255 & 4258 & 4256 & 4257 \\
 & hate  & 6250 & 1564 & 1561 & 1563 & 1562 \\
\hline
\end{tabular}
\label{table3}
\end{table}

\begin{table*}[htp]
\caption{Examples of the HateDebias dataset.}
\centering
\renewcommand\arraystretch{1}
\resizebox{\textwidth}{!}{
\begin{tabular}{c|c|l}
\hline
\begin{tabular}[c]{@{}c@{}}\textbf{Sensitive}\\ \textbf{Attribute}\end{tabular} & \multicolumn{1}{c|}{\textbf{Value}}       & \multicolumn{1}{c}{\textbf{Text}}                                                                                                         \\ \hline
                            & elder     & \cellcolor{lightpurple!60}\begin{tabular}[c]{@{}l@{}}user user royal calcutta was established in1829 and isnt in top 10 oldest golf clubs. it is the oldest\\ outside britain tho.\end{tabular}                                                                          \\ \cline{2-3} 
                            & elder     & \cellcolor{lightpurple!60}and then there was one hashtag i think the old man carl theory is pretty sick all he has left is the magnum…                                                                                      \\ \cline{2-3} 
                            & median   & \cellcolor{lightpurple!60}user and it reminds me too much of parties i had with friends when i was younger . minus the cam, lol.                                                                                           \\ \cline{2-3} 
\multirow{-4}{*}{Age}       & median   & \cellcolor{lightpurple!60}"ah feminism is still going strong, " "at least we\'re still two young, hot, blondes." " hashtag"                                                                                                 \\ \hline
                            & non-US & \cellcolor{lightgreen!60}how one woman used hashtag to reclaim her muslim american identity url via user url                                                                      \\ \cline{2-3} 
                            & non-US & \cellcolor{lightgreen!60}user user the only people that are stupid is you, vanderbilt did not build america, he owned or poorly paid…                                                                             \\ \cline{2-3} 
                            & US     & \cellcolor{lightgreen!60}user user that's only because all of the african american population came out to vote. you really do…                               \\ \cline{2-3} 
\multirow{-4}{*}{Country}   & US     & \cellcolor{lightgreen!60}\begin{tabular}[c]{@{}l@{}}latinos who think they're woke by choosing coke over pepsi and not knowing how bad coke has screwed latin \\ america hashtag\end{tabular}                                \\ \hline
                            & non-white       & \cellcolor{lightblue!40}what if my skin is so white it glows, and i've got blue hair? hashtag url                       \\ \cline{2-3} 
                            & non-white       & \cellcolor{lightblue!40}\begin{tabular}[c]{@{}l@{}}user there is literally nothing more creatively pathetic than taking an existing film and just making a \\ black or female version.\end{tabular}    \\ \cline{2-3} 
                            & white       & \cellcolor{lightblue!40}\begin{tabular}[c]{@{}l@{}}cause it don't pay to inform your own, it's more profitable to point the finger at the white man all the time \\ and not take any ownership.\end{tabular}     \\ \cline{2-3} 
\multirow{-4}{*}{Ethnicity} & white       & \cellcolor{lightblue!40}\begin{tabular}[c]{@{}l@{}}user i haven't seen anything other than clips and then read a bunch of reviews from black people about \\ why they hated it\end{tabular}                                   \\ \hline
                            & male        & \cellcolor{lightpink!60}user user user user from left to right, it's like the steps of male pattern hair loss.         \\ \cline{2-3} 
                            & male        & \cellcolor{lightpink!60}\begin{tabular}[c]{@{}l@{}}hashtag bunch of 1st world feminists dictating to a third one country in sexist \\ racism way how a male positive ht is bad for women\end{tabular}  \\ \cline{2-3} 
                            & female      & \cellcolor{lightpink!60}\begin{tabular}[c]{@{}l@{}}how many top grossing films with a strong female lead do we have to have before hollywood acknowledges \\ a winning strategy ?\end{tabular}                   \\ \cline{2-3} 
\multirow{-4}{*}{Gender}    & female      & \cellcolor{lightpink!60}rt user : user 2006 self made man: norah vincent chooses female privilege over male privilege url                                                                                                      \\ \hline
\end{tabular}}
\label{table14}
\end{table*}

\paragraph{Diversity} 

Following the Huang et al. \cite{huang-etal-2020-multilingual}, we select the English text from the hate speech detection task corpus from the Twitter platform \cite{huang-etal-2020-multilingual} to construct HateDebias. These texts are annotated with hate speech labels \cite{waseem-hovy-2016-hateful, Founta} via the Twitter streaming API\footnote{https://developer.twitter.com/}, along with the corresponding user profiles by analyzing their profile images via Face++\footnote{https://www.faceplusplus.com/}. {We guarantee the diversity of bias attributes since this corpus contains four types of demographic attributes, including age, country, gender, and ethnicity. Specifically, the protected groups based on age are the elderly and median, the gender-based protected groups encompass males and females, the protected groups based on country involve US and non-US, and the ethnicity-based protected groups are white and non-white. In this dataset, the validation set consists of 12681 samples, and the test set comprises 12682 samples.} 

\paragraph{Variability} 
To achieve the variability of bias types in dataset $\mathcal{D}$, our strategy is to generate sub-datasets with various types of bias and rearrange the sequence of these sub-datasets. We first collect a training set of 23,276 samples from the original training set, which consists of 59,179 samples, by removing samples with missing values, specifically those missing demographic attributes. This is because the continuous debiasing task requires the known bias attribute for its corresponding dataset. Based on the collected dataset with four bias attributes, we further partition it into four distinct subsets. To ensure distribution consistent with the collected data, the key issue in this partitioning is maintaining the same proportion of samples in each subset as in the overall dataset \cite{10.5555/1643031.1643047}. We address this issue using an equal proportion division method. Specifically, we randomly stratify the entire dataset based on a combination of labels (hate speech and non-hate speech) and demographic attributes (age, gender, country, and ethnicity), constraining (i) the proportion of each attribute in various sub-dataset is the same as the completed data, and (ii) there are not overlap sample between different sub-dataset. The results are shown in Table \ref{table3}, there are four subsets, `Age', `Gender', `Country', and `Ethnicity'. One can see that the proportion of each attribute among them is consistent with the completed data. To simulate the scenarios with continuous and changing bias types, we first assume only one explicit bias attribute in the sub-dataset and no other bias attribute can be observed. It means that, for any subset $D_i$, only the bias attribute $A_i$ is under-investigated. Furthermore, we randomly generate the sequence of these sub-datasets such that for the completed dataset $\mathcal{D}$ with its partition $\{D_1, D_2, D_3, D_4\}$, there exists a significant changing of bias attribute.

\paragraph{Example} We provide sample examples from the HateDebias Dataset in Table ~\ref{table14}, with one sample per row. We randomly selected four examples from subsets containing 5819 samples, covering both biased and unbiased cases for each sensitive attribute. Each color sequence represents specific bias variations in the HateDebias dataset, reflecting real-world biases. The dataset includes various sensitive attributes, such as age, country, gender, and ethnicity, converted into binary format. Table ~\ref{table3} details the number of samples and values for each attribute. The text field contains examples from a Twitter dataset for hate speech classification, with sensitive information masked to protect privacy. The symbol \ding{52} indicates bias for the corresponding attribute, while \ding{56} indicates no bias.

\subsection{Continual-learning-based Framework}\label{s6}

\begin{figure*}
  \centering
  \includegraphics[width=0.8\textwidth]{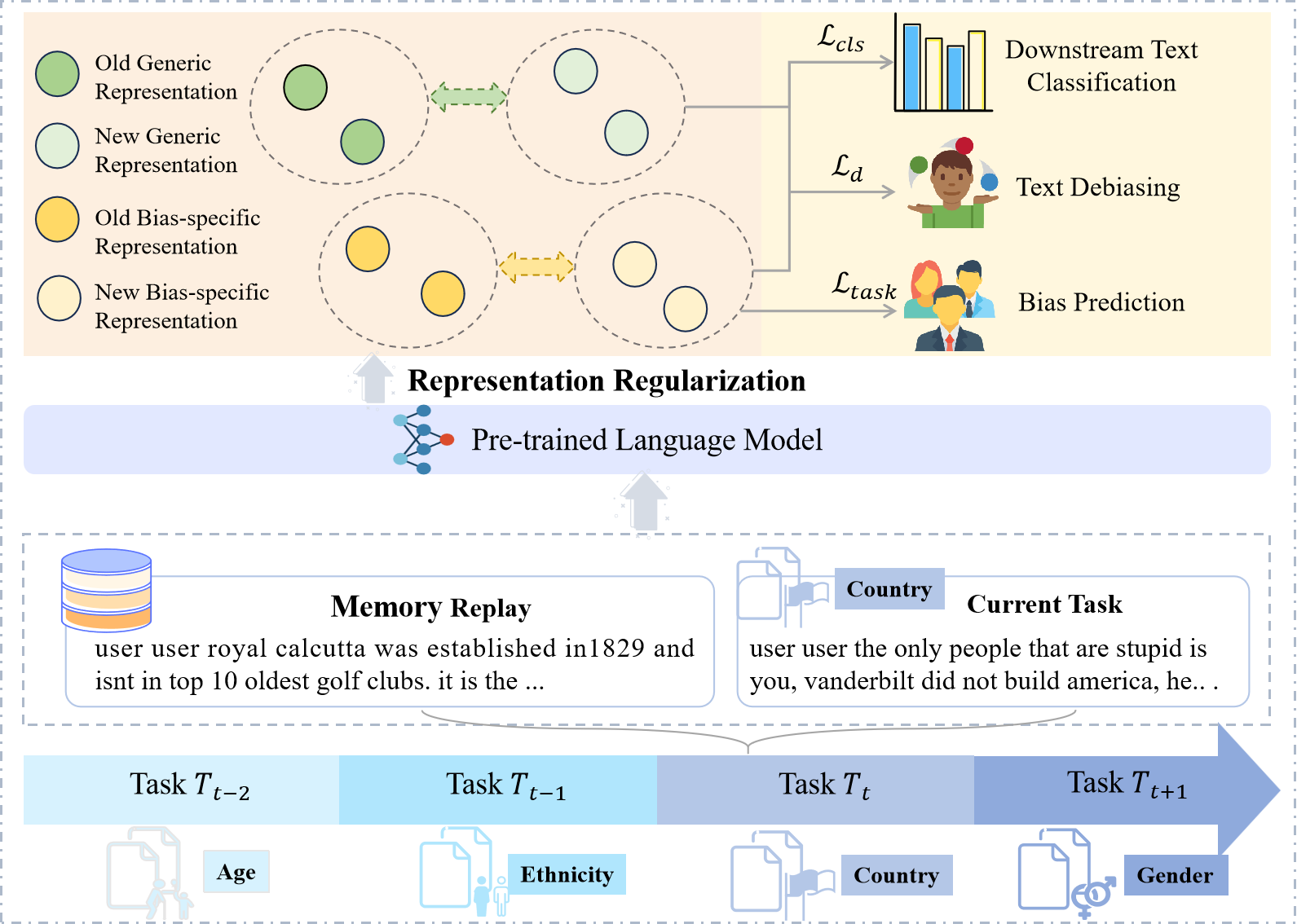}
  \caption{Continual-learning-based Framework. \textbf{Memory replay} selects samples from prior tasks to solidify previously acquired knowledge. \textbf{Bias information regularization} maintains shared knowledge across tasks and integrate bias-specific information.} 
  \label{fig:2} 
\end{figure*}

To further understand the continuous debiase task and verify the benchmark dataset, we develop a \textbf{continual-learning-based framework (CLF)} for continuous debiasing, aimed at improving model performance and fairness by learning attribute-irrelevant features to reduce bias. Our method uses a two-stage training process: the initial training stage and the memory replay stage. Initially, the model trains on current task data to acquire new knowledge. In the memory replay stage, representative samples from previous tasks are selected to update the memory for consolidating prior knowledge. To preserve shared knowledge across tasks while integrating bias-specific knowledge, we propose a bias information regularization technique to separate bias information into a generic space and a bias-specific space, facilitating interaction and helping the model adapt to various debiasing tasks. The pseudocode of the implementation process of our proposed continual-learning-based debiasing framework is shown in Algorithm \ref{algorithm:1}.

\begin{algorithm}[htp]
\small
\caption{CLF}
\label{algorithm:1}
\begin{algorithmic}[1]
\Require A set of training task $\{{T_i}^n\}$, the corresponding data set $\{{D_i}^n\}$, the store
ratio $\gamma$. Initial pre-trained language model $LM^0$.

\Ensure Trained language model encoder $LM^n$
\State Load pre-trained language model $LM^0$
\For{$i = 1,..., n$}
\State Calculate $\mathcal{L}_{cls}$ with $LM^i$ (Eq. \ref{lcls}) \Comment{Text Classification}
\State Calculate $\mathcal{L}_{d}$ with $LM^i$ (Eq. \ref{ld}) \Comment{Text Debiasing}
\If{$i \ge 2$}
\State Calculate $\mathcal{L}_g$ and $\mathcal{L}_s$ with $LM^i$ (Eq. \ref{lg}-\ref{ls}) \Comment{Bias Information Regularization}
\State Calculate $\mathcal{L}_{task}$ with $LM^i$ (Eq. \ref{ltask}) \Comment{Task Identification}
\State $M_t$ = []
\For{$r = 1,...,i$} \Comment{Memory Replay}
\State $M_t \leftarrow \gamma |D_r|$ (Eq. \ref{mtr})
\State $M_t \leftarrow M_t \cup M_t^r$
\EndFor
\EndIf
\State Calculate overall objective $\mathcal{L}$ with $LM^i$ (Eq. \ref{loverall}) \Comment{Overall Objective}
\State Update model parameters with $\mathcal{L}$
\EndFor
\end{algorithmic}
\end{algorithm}

\subsubsection{Overview}

We present a continual-learning-based framework for continual debiasing tasks, aiming to enhance the model's performance and fairness by promoting the learning of attribute-irrelevant features to effectively mitigate bias. For a given task sequence $\mathcal{T}_{\textit{debias}} = \{\operatorname{T_1}, \cdots, \operatorname{T_n}\}$, our proposed method is adopted to the two-stage training backbone, in terms of the \textit{initial training stage} and the \textit{memory replay stage}, where the replay-based method can effectively alleviate the classifier bias. During the \textit{initial training stage}, the model performs training solely on the current task data, allowing the learning of new task-related knowledge. Subsequently, in the \textit{memory replay stage}, a subset consisting of representative samples from the previous task data is selected to update the memory. The model is further retrained using the balanced memory bank, facilitating the retrieval and consolidation of previously acquired knowledge. Specifically, after training on $(t-1)$-th $(t \geq 2)$ task $\operatorname{T}_{t-1}$, $\gamma |D_{t-1}|$ examples are sampled from the training set $D_{t-1}$ into the memory buffer $M$, where $0 \leq \gamma \leq 1$ is the store ratio. Data from $M$ is then merged with the next training set $D_t$ when conducting the learning of task $t$. What's more, in the process of continual learning, an important aspect arises from the presence of shared knowledge across different debiasing tasks. Hence, the model must constantly adapt to the specific knowledge related to each debiasing task while avoiding forgetting the old knowledge throughout the learning process. To tackle this challenge, we disentangle sensitive attribute information from raw text representation to obtain a generic space and bias-specific space, introducing an \textit{bias information regularization technique} to allow interaction between two spaces. The regularization approach aims to preserve shared knowledge while enabling the model to effectively incorporate bias-specific knowledge for different debiasing tasks.

\subsubsection{Bias Information Regularization}
The operation of disentangled enables the separation of generic and bias-specific information in the hidden representations, facilitating a more refined analysis and understanding of the sentence’s underlying structure and context. The disentangled concept we define allows the model to learn generic information about downstream tasks and specific information about sensitive attributes, respectively. {For a given sentence $x$, a pre-trained language model $LM(\cdot)$ is employed to derive the hidden representations $h$, encompassing both generic and bias-specific information.} Subsequently, we design two disentangled networks, represented by $G(\cdot)$ and $S(\cdot)$, which play a vital role in extracting the generic representation $g$ and sensitive bias-specific representation $s$ from the previously obtained hidden representations $h$. $G(\cdot)$ and $S(\cdot)$ refer to two distinct feed-forward networks. The dimensions of the representations produced by $G(\cdot)$ and $S(\cdot)$ are identical to the hidden representations $h$. This systematic process enables the generic and bias-specific information to disentangle, facilitating a more refined analysis and understanding of the sentence's underlying structure and context. For the dynamic debiasing task of sensitive attributes, we train a classifier by exploiting information from two representations and allowing different representation spaces to be changed to different extents, maintaining the ability of the models when debiasing the sensitive attributes.

\paragraph{Generic Representation}
The generic representation $g$ is derived from the hidden representation feed-forward network $G(\cdot)$ that contains information common to various debiasing tasks within a given task sequence to alleviate forgetting old knowledge. When transitioning from one debiasing task to another, it is expected that the generic information remains relatively consistent across different tasks. A generic representation is used to learn the ability of downstream tasks and the ability of bias mitigation tasks.

\paragraph{Bias-specific Representation}\label{bsr}
Models require specific attribute information for effective performance in debiasing tasks. It is crucial for models to accurately identify the specific type of bias present within each text to facilitate effective debiasing. Consequently, we introduce a straightforward bias-identifier prediction task targeting the bias-specific representation $s$, which entails the classification of a given example into its corresponding task category. By adopting this auxiliary approach, we promote models' learning of distinct bias-specific information within $s$. The loss function for the bias-identifier predictor, denoted as $f_{task}$, is formulated as follows: 
\begin{equation}\label{ltask}
\mathcal{L}_{task} = E_{(x,z) \in D_t \cup M_t} \mathcal{L}_{CE} (f_{task}(S(LM(x)), z),
\end{equation}

where $z$ is the corresponding debiasing task id for $x$, $L_{CE} (\cdot)$ is the cross entropy loss function and $LM(\cdot)$ is the pre-trained model for text encoding in our framework. The memory buffer, denoted as $M_t$, retains samples from the training sets of prior tasks. This auxiliary setup serves as a valuable mechanism for enhancing task-specific learning and contributes to the robustness and adaptability of the model across different tasks.

\paragraph{Representation Regularization}

To cope with catastrophic forget during continually debiasing pre-trained language models via continual learning, we propose a regularization approach applied to both generic representations $g$ and bias-specific representations $s$, which preserves knowledge in distinct spaces separately to encourage positive transfer and mitigate forgetting. 

{We design two disentangled networks $G^{t-1}(\cdot)$ and $S^{t-1}(\cdot)$. We compute the generic representation $g$ via $G^{t-1}(LM^{t-1}(x))$ and the bias-specific representation $s$ via $S^{t-1}(LM^{t-1}(x))$ based on $D_{t-1}$, thereby preserving generic knowledge from the prior models to alleviate forgetting. 
Then, for the debiasing task $t$, we obtain the generic representations and bias-specific representations for all sentences within the training set $D_{t}$ and the memory buffer $M_{t}$ ($M_t \subseteq M$).
Finally, the general and bias-specific representations from task t-1 are retained together with the representations obtained in task $t$ for L2 norm regularization, thereby limiting the extent of updates to the general representation space and bias-specific representation space. Thus, during the learning for training pairs in the debiasing task $t$, we impose two regularization losses respectively:
\begin{equation}\label{lg}
\small
\mathcal{L}_g = E_{x \in D_t \cup M_t} || G^{t-1}(LM^{t-1}(x)) - G^{t}(LM^{t}(x)) ||_2,
\end{equation}
\begin{equation}\label{ls}
\small
\mathcal{L}_s = E_{x \in D_t \cup M_t} || S^{t-1}(LM^{t-1}(x)) - S^{t}(LM^{t}(x)) ||_2, 
\end{equation}
}
where $||\cdot||_2$ is the L2 norm regularization operation.

\subsubsection{Text Classification}\label{tc}
When performing the $t$-th debiasing task, we leverage a fusion of the generic representation $g$ and the bias-specific representation $s$ to facilitate text classification. This combination enables the model to effectively capture both the generic and bias-specific information in the downstream task. To optimize model performance, we minimize the cross-entropy loss:
\begin{equation}\label{lcls}
    \mathcal{L}_{cls} = E_{(x,y) \in D_t \cup M_t} \mathcal{L}_{CE} (f_{cls}(g \oplus s), y)),
\end{equation}
where $y$ denotes the corresponding class label assigned to input $x$, and $f_{cls}(\cdot)$ represents the class predictor function. The symbol ``$\oplus$" signifies the concatenation operation applied to the two representations under consideration.

\subsubsection{Text Debiasing}\label{td}
In the $t$-th debiasing stage, the model is aimed at debiasing the sensitive attribute $A_t$. Specifically, for a given sentence $x$, we adopt the given debiaser $H_t(\cdot)$ to mitigate sensitive attribute $A_t$ in the text. The debiaser $H_t(\cdot)$ is designed to be highly flexible and extensible, allowing for modifications as needed. In our experiments, we utilize four debiasing methods as the backbones of the debiaser $H_t(\cdot)$, namely FGM \cite{miyato2016adversarial}, PGD \cite{madry2017towards}, ATC \cite{han-etal-2021-diverse} and CL \cite{NEURIPS2020_d89a66c7}. {More details are presented in Section \ref{app:Debiasing Methods}.} We minimize the debiaser's loss:
\begin{equation}\label{ld}
    \mathcal{L}_d= E_{(x,y) \in D_t \cup M_t} \mathcal{L}_d(H_t(x)).
\end{equation}

We denote $\mathcal{L}_d(\cdot)$ as the loss function associated with the debiaser, responsible for executing the debiasing operation on the given sample.

\subsubsection{Memory Replay}

Since we only store a small number of samples from the training set of the previous debiasing task to achieve balance replay with the additional memory cost and training time, we need to choose the samples carefully so that the memory buffer $M$ is used efficiently. The previous memory replay methods \cite{riemer2018learning,wang-etal-2019-sentence} in continual learning solely focus on recalling representative samples to maintain the performance of fine-tuned downstream tasks. However, these samples may not necessarily be biased and thus may not be suitable for attribute-incremental continual learning methods. To address this limitation, We explore strategies to selectively recall specific samples in order to mitigate model bias.

Based on the task's definition, it becomes apparent that bias emerges due to variations in the model's performance among distinct groups of samples characterized by a sensitive attribute. Therefore, recalling samples with a low accuracy of labels under the current model for replaying can further alleviate the bias of the model, because when there are a large number of samples with low accuracy in the subset of the sensitive attributes, the model will have a greater bias for a sensitive attribute.

In the $t$-th debiasing stage, before fine-tuning and debiasing the model, we select the corpus $D_r$ for stage $r (r \textless t)$. First, the class predictor $f_{cls}(\cdot)$ is used to predict $D_r$, and the probability list $P$ corresponding to the true label of each sample is obtained. We set the number of retained samples as $\gamma |D_r|$, and obtain the corresponding probability threshold as $\beta$, then the samples retained in stage $r$ are:
\begin{equation}\label{mtr}
    M_t^r = \{ x \in D_r | P(x) \textless \beta \} (M_t^r \subseteq M_t).
\end{equation}

\subsubsection{Overall Objective}
The final loss of our dynamic debiasing model is:
\begin{equation}\label{loverall}
    \mathcal{L} = \mathcal{L}_{cls} + \alpha \cdot \mathcal{L}_d + \sigma \cdot (\mathcal{L}_g + \mathcal{L}_s + \mathcal{L}_{task}).
\end{equation}

We set the coefficient of the first loss terms to 1 for simplicity and only introduce two loss weights to tune: $\sigma$ and $\alpha$. The loss weight $\alpha$ represents the debiasing degree of the model, while the loss weight $\sigma$ controls the regularization degree of bias information.

\subsubsection{Generalizability of Our Framework}
Our framework for continuous debiasing is designed to be adaptable and extendable to various continuous learning methods. This includes replay strategies, regularization strategies, and other continual learning approaches. The primary advantage of our framework is its flexibility, allowing for the direct and convenient application of existing methods to the task of continuous debiasing. This adaptability ensures that the framework can be effectively employed for ongoing bias mitigation across diverse and evolving datasets, facilitating improved model fairness and robustness over time. As new techniques and strategies are developed, they can be integrated into our framework, providing a robust and evolving solution for dynamic debiasing challenges.

\section{Experiment}
We design multiple sets of experiments to validate the effectiveness of our proposed model on the proposed dataset. First, we evaluate that existing methods can hardly achieve the ideal performance in the multiple biases while the proposed method can solve this problem, which is shown in \ref{exp1} and Sec. \ref{exp2}. Sequentially, to further evaluate that our method can achieve stable performance under the continuous scenarios of our datasets, we further consider different numbers of biases in the continuous scenarios, which is shown in Sec. \ref{exp3}. Moreover, we evaluate that our method can avoid catastrophic forgetting as shown in Sec. \ref{exp4}. Besides, we conduct ablation experiments, parameter search experiments, statistical significance tests, explore the effects of different sequence orders, and different pre-trained language models.

\subsection{Setups}
\label{set up}

\paragraph{Debiasing Methods}
\label{app:Debiasing Methods}
{We conduct experiments based on four debiasing methods in the continuous debiasing task. 
\textbf{Fast Gradient Method (FGM) \cite{miyato2016adversarial}} follows the adversarial training method to add perturbation to debiasing model representation module. 
\textbf{Projected Gradient Descent (PGD)} \cite{madry2017towards} is another unsupervised adversarial learning method, which adds an iterative attack that differs from FGM. 
\textbf{Adversarial Training Classifiers (ATC)} tend to perform supervised adversarial training \cite{han-etal-2021-diverse, kashyap2022towards, Clavijo_2022} to predict sensitive attributes for adversarial learning, thereby blurring the model's identification of sensitive attributes.
\textbf{Contrastive Learning (CL) \cite{NEURIPS2020_d89a66c7}} aims at pulling identical entailment pairs along opposite sensitive attribute directions closer, which serves as an intermediate pre-training approach to mitigate biases in contextual representations.} 

\paragraph{Baselines}
\label{app:baseline}
{We compare with the following baselines to verify the effectiveness of our proposed method.
\textbf{Fine-tune \cite{DBLP:journals/corr/abs-1901-11373}} modifies the parameters of the pre-trained language model sequentially.
\textbf{Experience Replay (ER) \cite{wang-etal-2019-sentence,riemer2018learning}} is a memory-based technique that involves retaining a limited subset of samples from prior tasks and subsequently recalling them during training.
\textbf{IDBR \cite{huang-etal-2021-continual}} incorporates information disentanglement regularization into the encoding process.
\textbf{Multi-task Learning (MTL) \cite{10.5555/3495724.3496213}} trains the model on all tasks simultaneously to avoid the problem of catastrophic forgetting, as a static debiasing method. 
Two-level Knowledge Distillation (TKD) \textbf{\cite{yang2023continual}} is a technique designed to improve the continual text classification. It involves transferring knowledge at the feature level and the output level.  
\textbf{Representation Projection Invariance (REPINA) \cite{razdaibiedina-etal-2023-representation}} aims to preserve the information content of representations by discouraging undesirable changes.  
\textbf{Instance-wise Relation Distillation (IRD) \cite{luo2023mitigating}} is a regularization technique used in continual learning to mitigate the forgetting of previously learned knowledge when the model learns new tasks.  
Since our experimental setting involves attribute-incremental continual learning, we transform the attribute-incremental setting into a debiasing-task-incremental setting to enable the task-incremental continual learning method to work on our dataset. }

\paragraph{Model Details}

\begin{table}
\caption{The optimal hyperparameter values of each debiasing method.}
\centering
\begin{tabular}{cccc}
\hline
Method & $\gamma$ & $\alpha$ & $\sigma$ \\
\hline
FGM & 0.1 & 1 & 0.1 \\
PGD & 0.1 & 1 & 0.05 \\
ATC & 0.05 & 0.1 & 0.05 \\
CL & 0.1 & 0.1 & 0.1 \\
\hline
\end{tabular}
\label{table4}
\end{table}

\begin{table*}[htp]
\setlength{\belowcaptionskip}{0.3cm}
\tabcolsep=0.06cm
\caption{Main results. The columns and rows are the evaluation metrics and the names of each method respectively. Values in brackets are the variances of the metrics. For the static debiasing method, since there are no multiple sequence experiments, we denote the variance as 0.00. \textbf{MTL} represents a static debiasing approach that addresses multiple sensitive attributes concurrently. \textbf{Fine-tune} constitutes the most straightforward experimental setup for continual debiasing tasks. Furthermore, \textbf{ER}, \textbf{IDBR}, \textbf{REPINA}, \textbf{IRD} and \textbf{TKD} are prevalent methods in continuous learning for continual debiasing tasks.}
\centering
\begin{tabular}{c|c|cc|ccccc|c}
\hline
\multirow{2}{*}{\textbf{Method}}& \multirow{2}{*}{\textbf{\makecell[c]{Debiasing \\ Method}}}& \multicolumn{2}{c|}{\textbf{Performance}} & \multicolumn{5}{c|}{\textbf{Fairness}} & \multirow{2}{*}{\textbf{DTO}}\\ 
\cline{3-9}
& & \textbf{Acc $\uparrow$} & \textbf{F1-macro $\uparrow$} & \textbf{FPED(Gender) $\downarrow$} & \textbf{FPED(Age) $\downarrow$} & \textbf{FPED(Country) $\downarrow$} & \textbf{FPED(Ethnicity) $\downarrow$} & \textbf{AAB $\downarrow$} & \\
\hline
 \multirow{5}{*}{\makecell[c]{MTL \cite{10.5555/3495724.3496213}}} & No-debias & 85.90(0.00)  & 83.00(0.00)  & 4.77(0.00)  & 7.77(0.00)  & 5.35(0.00)  & 4.43(0.00)  & 5.58(0.00) & {0.28}\\
& PGD & 86.79(0.00)  & 83.59(0.00)  & \textbf{3.05(0.00)}  & 7.37(0.00)  & \textbf{3.21(0.00)}  & \textbf{3.87(0.00)}  & \textbf{4.37(0.00)} & {\textbf{0.02}}\\
& FGM & 87.19(0.00)  & 83.99(0.00)  & 3.92(0.00)  & \textbf{6.05(0.00)}  & 4.97(0.00)  & 4.18(0.00)  & 4.78(0.00) & {0.10}\\
& ATC & 85.48(0.00)  & 82.32(0.00)  & 6.11(0.00)  & 8.28(0.00)  & 5.55(0.00)  & 3.87(0.00)  & 5.95(0.00) & {0.37}\\
& CL & \textbf{88.12(0.00)}  & \textbf{84.38(0.00)}  & 4.76(0.00)  & 7.04(0.00)  & 6.26(0.00)  & 5.17(0.00)  & 5.81(0.00) & {0.33}\\
\hline
\multirow{5}{*}{\makecell[c]{Fine-tune \cite{DBLP:journals/corr/abs-1901-11373}}} & No-debias & 86.73(2.92)  & 83.51(3.23)  & 5.55(0.91)  & 8.67(2.16)  & 5.97(1.67)  & 4.51(1.10)  & 6.17(0.65) & {0.42}\\
& PGD & 87.14(3.11)  & 84.23(3.42)  & 5.31(0.68)  & 8.67(4.52)  & 5.56(0.68)  & 4.06(0.60)  & 5.90(0.75) & {0.35}\\ 
& FGM & 87.07(3.31)  & 84.16(3.68)  & 5.43(1.22)  & 8.86(5.80)  & 5.74(0.93)  & 4.39(0.52)  & 6.10(1.01) & {0.40}\\
& ATC & 88.14(1.34)  & 85.17(1.36)  & 5.95(0.60)  & 10.02(2.00)  & 6.56(1.19)  & 4.85(0.88)  & 6.84(0.30) & {0.57}\\
& CL & 86.59(2.72)  & 83.41(3.32)  & 5.15(0.89)  & 8.85(1.70)  & 6.19(0.87)  & 4.28(0.83)  & 6.12(0.37) & {0.40}\\
\hline
\multirow{4}{*}{\makecell[c]{ER \cite{wang-etal-2019-sentence}}} & PGD & 88.68(0.32)  & 85.64(0.37)  & 5.38(0.70)  & 8.63(0.70)  & 5.43(0.86)  & 4.64(0.82)  & 6.02(0.31) & {0.38}\\
& FGM & 88.64(0.55)  & 85.60(0.68)  & 5.53(0.23)  & 9.04(0.74)  & 5.27(0.42)  & 4.83(0.87)  & 6.17(0.21) & {0.42}\\
& ATC & 87.17(1.72)  & 83.83(1.61)  & 5.10(0.62)  & 8.58(2.41)  & 6.32(1.52)  & 4.79(1.13)  & 6.20(0.37) & {0.42}\\
& CL & 87.19(0.73)  & 84.05(0.88)  & 4.96(0.93)  & 8.99(2.43)  & 6.25(0.82)  & 4.83(1.59)  & 6.26(0.47) & {0.44}\\
\hline
\multirow{2}{*}{\makecell[c]{IDBR \cite{huang-etal-2021-continual}}} & ATC & 88.74(0.87)  & 85.81(1.40)  & 5.30(1.34)  & 9.45(1.99)  & 6.53(1.30)  & 4.76(0.48)  & 6.51(0.51) & {0.49}\\
& CL & 87.57(0.80)  & 84.40(0.90)  & 4.33(1.18)  & 8.18(1.43)  & 5.32(1.24)  & 4.70(1.21)  & 5.63(0.64) & {0.29}\\
\hline
\multirow{2}{*}{\makecell[c]{REPINA 
 \cite{razdaibiedina-etal-2023-representation}}} & ATC & 87.42(0.97) & 84.27(1.21) & 5.71(0.79) & 9.72(3.06) & 5.83(0.45) & 4.56(1.00) & 6.46(0.37) & {0.48}\\
& CL & 88.50(0.84) & 85.29(1.31) & 6.05(1.52) & 9.73(4.36) & 6.64(1.40) & 4.40(1.21) & 6.70(1.12) & {0.54}\\
\hline
\multirow{2}{*}{\makecell[c]{IRD \cite{luo2023mitigating}}} & ATC & 88.70(0.77) & 85.79(1.00) & 5.66(0.70) & 9.94(1.37) & 6.74(1.10) & 5.28(0.62) & 6.90(0.45) & {0.58}\\
& CL & 87.11(0.97) & 83.84(1.19) & 4.41(2.18) & 8.02(3.92) & 5.40(1.37) & 4.49(1.65) & 5.58(1.14) & {0.28}\\
\hline
\multirow{2}{*}{\makecell[c]{TKD \cite{yang2023continual}}} & ATC & \textbf{88.86(1.07)} & \textbf{85.96(1.34)} & 5.58(0.79) & 9.66(0.87) & 6.37(1.23) & 4.81(1.07) & 6.60(0.26) & {0.51}\\
& CL & 87.02(1.51) & 83.70(1.61) & 4.58(1.86) & 8.42(2.00) & 5.33(0.05) & 4.51(1.30) & 5.71(0.91) & {0.31}\\
\hline
\multirow{4}{*}{CLF} & PGD & 87.16(2.02)  & 84.01(2.16)  & 5.79(0.93)  & 8.64(1.48)  & 4.96(0.56)  & 4.38(0.89)  & 5.94(0.46) & {0.36}\\
& FGM & 86.37(7.63)  & 83.35(7.12)  & 5.78(0.93)  & 8.00(4.16)  & 4.57(0.52)  & 4.35(2.65)  & 5.67(0.70) & {0.30}\\
& ATC & 83.85(13.81)  & 80.51(13.78)  & 3.90(2.51)  & 7.33(5.08)  & 4.06(2.82)  & \textbf{4.22(1.26)}  & 4.88(1.28) & {0.12}\\
& CL & 83.94(6.91)  & 80.31(6.42)  & \textbf{2.79(2.82)}  & \textbf{6.21(2.60)}  & \textbf{3.61(2.65)}  & 4.84(1.60)  & \textbf{4.36(1.45)} & {\textbf{0.06}}\\
\hline
\end{tabular}
\label{table5}
\end{table*}

We conduct all experiments based on PyTorch and an A30 with 24 GB of memory. We design our framework based on Transformers. Furthermore, we choose the bert-base-uncased model \cite{devlin-etal-2019-bert} as the backbone of our framework. We set the same hyper-parameters with a fixed initialization seed for our model's training, where the batch size is 32 and the feature dimension is 768. The max length of the input samples is 32. The BERT encoder with a learning rate of 5e-5 is optimized utilizing Adam. For the MTL model, we set the training epoch to 10, while for other continual learning methods, we set the number of epochs to 5 for each stage. We perform a grid search on the hyperparameters in the model. The optimal hyperparameter values of each debiasing method are shown in Table \ref{table4}. In order to avoid the contingency caused by the order of debiased attributes, we conduct experiments with $A_4^4=24$ combinations of 4 attributes. 

\paragraph{Evaluation Metrics}


{Unlike traditional task-incremental continual learning tasks, we perform attribute-incremental continual learning on the hate speech detection task to debias for a specific attribute. We adopt the accuracy (Acc) and macro-average $F_1$ (F1-macro) value to evaluate the performance of models on the hate speech detection task. To evaluate model fairness, we measure the equality differences of false positive rates (FPED) across different groups for a specific attribute. As formulated in Eq. \ref{fped}, FPED involves aggregating the differences between the false positive rates of each subgroup $FP R_d$ and the overall false positive rate $FPR$ within a demographic factor $D$. Since there are 24 sequence settings for each attribute in the attribute-incremental continual learning task, we calculate the average FPED across all task sequences for each attribute in our continual-learning-based framework. In terms of the overall fairness of the model, we compute the average FPED of four attribute biases to measure the average attribute bias (AAB). Since there are 24 sequence settings in the attribute-incremental continual learning task, we calculate the AAB for all task sequences for our continual-learning-based framework. For each metric, we also calculate the variance.  Moreover, we further introduce the DTO metric \cite{han-etal-2022-balancing,han-etal-2022-fairlib} to measure the trade-off between the fairness improvements and performance impacts. A lower DTO value indicates a more effective trade-off.}


{
\begin{equation}
\label{fped}
F P E D=\sum_{d \in D}\left|F P R_{d}-F P R\right|.
\end{equation}
}
\subsection{Debiasing Performance Forgetting of Continuous Debias-
ing}\label{exp1}

{As shown in Table \ref{table5}, we analyze the results to explore the impact of persistent bias on the fairness of models and whether the debiasing performance of the model suffers from the catastrophic forgetting problem in the continuous debiasing task.
Compared MTL-based methods representing static debiasing methods with Fine-tune-based methods indicating continuous debiasing methods, the results reveal that the AAB of the model under continuous debiasing tasks is consistently higher than that of static debiasing setting, demonstrating stronger bias. This indicates that the continuous debiasing performance of models may suffer from catastrophic forgetting with ever-changing biases.} 
The HateDebias dataset closely mirrors real-world bias occurrences, validating that static methods are inadequate under continual biases. Thus, it is significant to develop adaptive debiasing methods that enhance the overall performance and fairness of models in practical applications.
This finding underscores the rationale and necessity of the HateDebias dataset proposed in this paper. It closely resembles the occurrence of bias in real-world scenarios, simulating multiple, continuously changing biases, and provides a real environment closer to practical applications.

\subsection{Performance of Different Methods on HateDebias}\label{exp2}

{As detailed in Table \ref{table5}, our experimental findings indicate that our approach yields substantial improvements for the fairness of models in continual debiasing tasks.
Static debiasing methods (MTL), which can be considered as an upper-bound followed by \cite{huang-etal-2021-continual,luo2023mitigating}, show significant debiasing effects when dealing with multiple sensitive attributes, while fine-tuning-based continuous debiasing methods have limited debiasing capabilities. The models with fine-tuning-based methods are less stable in debiasing and less effective in addressing continuous bias, with larger AAB values and the variances.} 
Additionally, the continuous debiasing effect using the ER, REPINA and IRD methods is worse, indicating an inability to effectively reduce bias. 
These methods are ineffective because the HateDebias dataset closely mirrors real-world bias occurrences, simulating multiple and continuously changing biases. Experimental results show that existing methods perform poorly in these complex, dynamic environments.
Finally, the experimental results show that our proposed framework significantly reduces bias in continuous debiasing tasks. Especially with supervised debiasing strategies, the model's bias is notably reduced at the cost of slight performance trade-offs in downstream tasks. Compared to methods like REPINA, IRD, and TKD, the CLF method demonstrates better debiasing effects and model performance across various debiasing strategies. {Our results demonstrate that CLF achieves the best trade-offs as measured by DTO, outperforming other methodologies included in the study. Although CLF shows lower raw performance in some metrics, it significantly improves fairness.}
Experimental results show the proposed method CLF excels in handling multiple and continuously changing biases in complex real-world scenarios, effectively debiasing while maintaining high performance and fairness.

\subsection{Performance of Debiasing Preservation Capability}\label{exp3}

To verify that our method can avoid catastrophic forgetting, we designed experiments to evaluate the debiasing preservation capability of the model in continuous learning tasks. We propose a Bias Change (BC) metric to assess the model's debiasing preservation capability:
\begin{equation}
\small
    BC = \frac{1}{n-1} \sum_{i=1}^{n-1} FPR_i^{n} - FPR_i^{i},
\end{equation}
where $FPR_i^{i}$ is the false positive rate of the $i$-th bias in the $i$-th stage. Table \ref{table6} presents the results of the debiasing preservation capability. 
These results demonstrate that our proposed CLF method significantly outperforms others in preserving debiasing effects over time. Specifically, the CLF method achieves a BC value of -1.60, indicating substantial bias reduction and showcasing its effectiveness in mitigating bias. In comparison, the REPINA method shows no improvement, while methods like TKD and IDBR exhibit moderate improvements.
The poor performance of the comparative methods indirectly reflects the reasonableness of our dataset in terms of capturing the diversity and variability of real-world biases.

\begin{table}
\centering
\caption{The results of debiasing preservation capability.}
\begin{tabular}{cccc}
\hline
\textbf{Method} & \textbf{BC} \\
\hline
CLF & -1.60(2.15) \\
IDBR & -0.61(1.46) \\
REPINA & 0.00(0.01)  \\
TKD & -0.82(1.24) \\
IRD & -1.10(1.67) \\
\hline
\end{tabular}
\label{table6}
\end{table}

\begin{table}
\centering
\tabcolsep=0.1cm
\caption{Experimental results of different sequence lengths.}
\centering
\begin{tabular}{c|c|cc|c}
\hline
\multirow{2}{*}{\textbf{\makecell[c]{Sequence \\ Length}}}& \multirow{2}{*}{\textbf{Method}}& \multicolumn{2}{c|}{\textbf{Performance}} & \multicolumn{1}{c}{\textbf{Fairness}} \\ 
\cline{3-5}
& & \textbf{Acc $\uparrow$} & \textbf{F1-macro $\uparrow$} & \textbf{AAB $\downarrow$ (Var.)}\\
\hline
\multirow{5}{*}{2} & IDBR & 87.51(3.67) & 84.52(4.12) & 6.37(2.30)\\
\cline{2-5}
& REPINA &  87.81(1.91) & 84.71(1.68) & 6.62(1.99) \\
\cline{2-5}
& IRD & 88.10(2.22) & 85.10(2.35) & 6.89(2.96)\\
\cline{2-5}
& TKD & \textbf{88.14(1.30)} & \textbf{85.31(1.28)} & 6.69(1.84)\\
\cline{2-5}
& CLF & 85.17(6.79)  & 81.91(7.13) & \textbf{5.77(2.23)} \\
\hline
\multirow{5}{*}{3} & IDBR & 87.96(1.28) & 84.90(1.52) & 6.09(0.96)\\
\cline{2-5}
& REPINA & \textbf{88.43(0.54)} & \textbf{85.28(0.84)} & 6.60(1.12) \\
\cline{2-5}
& IRD & 86.75(4.56) & 83.75(4.78) & 6.00(0.80)\\
\cline{2-5}
& TKD & 87.31(1.77) & 84.29(2.11) & 6.19(0.72)\\
\cline{2-5}
& CLF & 85.78(2.35)  & 81.81(3.23) & \textbf{4.96(1.55)}\\
\hline
\multirow{5}{*}{4} & IDBR & 87.57(0.80)  & 84.40(0.90)  & 5.63(0.64)\\
\cline{2-5}
& REPINA & \textbf{88.50(0.84)} & \textbf{85.29(1.31)} & 6.70(1.12)\\
\cline{2-5}
& IRD & 87.11(0.97) & 83.84(1.19) &5.58(1.14)\\
\cline{2-5}
& TKD & 87.02(1.51) & 83.70(1.61) & 5.71(0.91)\\
\cline{2-5}
& CLF & 83.94(6.91) & 80.31(6.42) & \textbf{4.36(1.45)} \\
\hline
\end{tabular}
\label{table13}
\end{table}

\begin{table*}[t]
\centering
\tabcolsep=0.13cm
\setlength{\belowcaptionskip}{0.3cm}
\caption{Ablation experiment results. Upon removal of the BIR module, both the performance and fairness metrics of the models deteriorated.}
\begin{tabular}{c|c|cc|ccccc}
\hline
\multirow{2}{*}{\textbf{\makecell[c]{Debiasing \\ Method}}}& \multirow{2}{*}{\textbf{Method}}& \multicolumn{2}{c|}{\textbf{Performance}} & \multicolumn{5}{c}{\textbf{Fairness}} \\ 
\cline{3-9}
& & \textbf{Acc $\uparrow$} & \textbf{F1-macro $\uparrow$} & \textbf{Gender $\downarrow$} & \textbf{Age $\downarrow$} & \textbf{Country $\downarrow$} & \textbf{Ethnicity $\downarrow$} & \textbf{AAB $\downarrow$} \\
\hline
\multirow{2}{*}{ATC} & \multicolumn{1}{l|}{CLF} & \textbf{83.85}  & 80.51  & \textbf{3.90}  & \textbf{7.33}  & \textbf{4.06}  & \textbf{4.22}  & \textbf{4.88} \\
& w/o BIR & 85.73  & \textbf{81.66}  & 4.42  & 8.67  & 5.17  & 4.57  & 5.71 \\ 
\hdashline
\multirow{2}{*}{CL} & \multicolumn{1}{l|}{CLF} & \textbf{83.94}  & \textbf{80.31}  & \textbf{2.79}  & \textbf{6.21}  & \textbf{3.61}  & 4.84  & \textbf{4.36} \\
& w/o BIR & 83.66  & \textbf{80.31}  & 4.78  & 7.44  & 4.81  & \textbf{4.24}  & 5.32 \\
\hline
\end{tabular}
\label{table7}
\end{table*}

\begin{table*}[t]
\setlength{\belowcaptionskip}{0.3cm}
\caption{Results of Deberta-v3-base and RoBERTa-base. The columns and rows of the table are the evaluation metrics and the names of each method respectively. Experimental results on different backbones show that our method has obvious continuous debiasing capabilities and is better than existing methods.}
\centering
\tabcolsep=0.03cm
\begin{tabular}{ccc|cc|ccccc}
\hline
\multirow{2}{*}{\textbf{\makecell[c]{Debiasing \\ Method}}} & \multirow{2}{*}{\textbf{Backbone}} & \multirow{2}{*}{\textbf{Method}} & \multicolumn{2}{c|}{\textbf{Performance}} & \multicolumn{5}{c}{\textbf{Fairness}} \\ 
\cline{4-10}
& & & \textbf{Acc $\uparrow$} & \textbf{F1-macro $\uparrow$} & \textbf{Gender $\downarrow$} & \textbf{Age $\downarrow$} & \textbf{Country $\downarrow$} & \textbf{Ethnicity $\downarrow$} & \textbf{AAB $\downarrow$} \\
\hline
\multirow{8}{*}{CL} & \multirow{4}{*}{\makecell[c]{Deberta \\ -v3-base}} & MTL & 87.94 & 84.90 & 06.01 & 10.07 & 4.91 & 4.97 & 6.49 \\
& & Fine-tune & 88.86 & 85.77 & 6.23 & 11.04 & 5.88 & 5.16 & 7.08 \\
& & IDBR & 86.15 & 82.88 & 5.63 & 8.98 & 4.37 & 5.46 & 6.11 \\
& & CLF & 83.95 & 80.85 & 5.35 & 9.24 & 4.56 & 4.54 & 5.92 \\
\cline{2-10}
& \multirow{4}{*}{\makecell[c]{RoBERTa \\ -base}} & MTL & 87.53 & 84.25 & 4.98 & 8.43 & 4.49 & 4.45 & 5.59 \\
& & Fine-tune & 88.16 & 85.07 & 5.17 & 9.07 & 4.47 & 4.62 & 5.83 \\
& & IDBR & 86.22 & 82.96 & 4.14 & 7.32 & 4.92 & 3.99 & 5.09 \\
& & CLF & 85.53 & 82.13 & 3.65 & 7.23 & 4.42 & 3.46 & 4.69 \\
\hline
\end{tabular}
\label{table9}
\end{table*}

\subsection{Effects of Different Length Settings on {HateDebias}}\label{exp4}

In practical applications, biases come in various types and forms. To gradually make our dataset approximate real-world scenarios, we experiment with different sequence lengths to evaluate the debiasing effects under different length settings. We compare continuous debiasing tasks with sequence lengths of 2, 3 and 4 to analyze how the diversity and variability of the dataset affect model performance and fairness.
As shown in Table ~\ref{table13}, we choose CL as the debiasing method and the results reveal several key insights: 
(1) As the sequence length increases, the performance metrics generally show a decreasing trend. This suggests that longer sequences may introduce more complexity, making it challenging for models to maintain high performance. (2) The AAB values decrease with longer sequence lengths, indicating improved fairness. This trend demonstrates that datasets with longer sequences better capture the diversity and variability of real-world biases, helping models to generalize and debias more effectively. (3) Our proposed method CLF exhibits the best continuous debiasing capability across different sequence lengths, indicating that it can effectively cover datasets in various real-world scenarios.

\subsection{Ablation Results}
{We discard the regularization losses $\mathcal{L}_g$ and $\mathcal{L}_s$ to conduct ablation experiments to verify the effectiveness of BIR, as shown in Table \ref{table7}.} It can be seen that when ablating BID, the AAB values of CLF with ATC and CLF with CL increased by 0.0083 and 0.0096, reflecting that the model is more biased. The result demonstrates that BIR can effectively alleviate the continual biases in the model. Simultaneously, BIR would affect the performance of the model in downstream tasks to a certain extent. When discarding BID, the accuracy of CLF with ATC and CLF with CL increased by 0.0188 and 0.0028.

We further explore the performance and fairness of the model utilizing our designed replay strategy and the ER strategy, respectively. It can be seen that the performance of BIR solely adopting our replay strategy in downstream tasks is not as good as those ER methods. Whereas from the perspective of the AAB metric for evaluating fairness, the replay strategy makes the model more fair. The experimental results show that the BIR strategy and our proposed memory replay strategy can effectively debias the pre-trained language model in the continuous debiasing task.

\subsection{Experiments on Other Pre-trained Models}

In addition to conducting experiments on the BERT model, we further perform our method with other pre-trained language models Deberta-v3-base \cite{he2023debertav} and RoBERTa-base \cite{DBLP:journals/corr/abs-1907-11692}. The experimental results are shown in the Table \ref{table9}. Experimental results verify the efficacy of our method in mitigating biases across various large-scale pre-trained models. This further validates the versatility and applicability of our approach in addressing bias within the realm of natural language processing models, underscoring its potential utility in enhancing the fairness and impartiality of machine learning applications.

\subsection{Statistical Significance Test}

\begin{table}[t]
\setlength{\belowcaptionskip}{0.3cm}
\centering
\caption{Results of Wilcoxon signed-rank test significance analysis. A \textit{p} value below 0.05 signifies that our method demonstrates statistically significant improvement over the comparison method.}
\begin{tabular}{cccc}
\hline
\textbf{Debiasing Method} & \textbf{Method} & \textbf{Compare Method} & \textbf{\textit{p}} \\
\hline
\multirow{3}{*}{ATC} & \multirow{3}{*}{CLF} & Fine-tune & 3.93e-06 \\
&  & Replay & 6.00e-02 \\
&  & IDBR & 6.56e-06 \\
 \hline
\multirow{3}{*}{CL} & \multirow{3}{*}{CLF} & Fine-tune& 6.04e-02 \\
&  & Replay& 5.13e-06 \\
&  & IDBR & 1.75e-04 \\
 \hline
\end{tabular}
\label{table12}
\end{table}

We have conducted a statistical significance test to validate the significance of the experiment results. We performed a Wilcoxon signed-rank test significance analysis \cite{woolson2007wilcoxon} on the experimental results of our model CLF (on 24 different debiasing sequences) as well as the experimental results of other continual debiasing methods. We chose ATC and CL as the debiasing methods for significance analysis. The experimental results are shown in Table \ref{table12}. The \textit{p} values of the significance analysis of our proposed model and the existing model are shown in the table. Experimental results show that our improvement is significant.

\section{Conclusion}
In this paper, we focus on the continuous debiasing task, addressing the problem of debiasing in hate speech scenarios with continuously varying bias types. We propose a dataset, HateDebias, to benchmark continuous debiasing tasks. The HateDebias dataset contains 23,276 samples with four bias attributes, corresponding to four sub-datasets. Based on HateDebias, we also propose a continuous-learning-based framework to eliminate emerging biases while preserving model performance in downstream tasks. We evaluate the debiasing performance of our methods and baselines on HateDebias, which demonstrates that our methods yield a promising performance for improving continuous debiasing tasks. 
{In the future, we will extend the CLF framework to handle intersecting bias attributes, which involves developing more sophisticated representation disentanglement mechanisms to isolate and debias multiple overlapping sensitive attributes simultaneously, as well as designing new evaluation metrics to assess fairness across intersecting groups. Additionally, we aim to explore how to avoid unintended bias amplification for non-targeted groups when debiasing intersecting attributes, further enhancing the framework’s applicability to complex fairness challenges.}

\section*{Acknowledgements}
Our work is supported by the National Social Science Fund of China (No. 22BTQ045).


\bibliographystyle{IEEEtran}
\bibliography{reb}








\end{document}